# Improve in-situ life prediction and classification performance by capturing both the present state and evolution rate of battery aging

Mingyuan Zhao, Yongzhi Zhang*

College of Mechanical and Vehicle Engineering, Chongqing University, 400030, Chongqing, China.

*Corresponding author: Yongzhi Zhang (yzzhangbit@gmail.com)

*Abstract*—This study develops a methodology by capturing both the battery aging state and degradation rate for improved life prediction performance. The aging state is indicated by six physical features of an equivalent circuit model that are extracted from the voltage relaxation data. And the degradation rate is captured by two features extracted from the differences between the voltage relaxation curves within a moving window (for life prediction), or the differences between the capacity vs. voltage curves at different cycles (for life classification). Two machine learning models, which are constructed based on Gaussian Processes, are used to describe the relationships between these physical features and battery lifetimes for the life prediction and classification, respectively. The methodology is validated with the aging data of 74 battery cells of three different types. Experimental results show that based on only 3-12 minutes' sampling data, the method with novel features predicts accurate battery lifetimes, with the prediction accuracy improved by up to 67.09% compared with the benchmark method. And the batteries are classified into three groups (long, medium, and short) with an overall accuracy larger than 90% based on only two adjacent cycles' information, enabling the highly efficient regrouping of retired batteries.

*Index Terms* — In-situ life prediction, triple-classification, physical features, machine learning, aging state indication, degradation rate extraction

## I. INTRODUCTION

Lithium-ion batteries have been widely used in transportation electrification, stationary energy storage, and portable electronics, *etc.* [1-3] The battery degradation in usage reduces its operation reliability, making the remaining useful life (RUL) prediction a vital function of the battery management system for safety concerns [4, 5]. However, the cell-to-cell variations of both the aging characteristics and working conditions not only make the RUL prediction a great challenge in real applications but also lead to high difficulties for the efficient classification of large-scale retired battery cells [6, 7].

The RUL prediction methods are mainly divided into model-based and data-driven methods. Model-based methods, on one hand, establish a parameterized function model to describe the performance decline trajectory of the battery based on experience [8, 9], and on the other hand, model the internal electrochemical reaction mechanisms of the battery from the bottom up to calculate the evolution of the internal states [10, 11]. However, the former requires a pre-specified function without considering any internal mechanism, and thus as the battery ages or the operating condition changes, the predicted decay trajectory through extrapolation could be inaccurate. The latter needs a balance between the physical modeling accuracy and the onboard computation capability, and faces challenges in both parameterization and validation [12]. Besides, many degradation mechanisms and their coupling effects are still poorly understood, so even the most advanced models have limited applicability for in-situ life prediction [13].

By contrast, data-driven methods do not need to understand the aging mechanisms, and instead utilize historical data to establish statistical or machine learning (ML) models to describe the relationships between input and output. Many ML methods have been developed for battery RUL prediction, with the typical methods including the relevance vector machine [14], Gaussian process regression (GPR) [15], and convolutional neural network [16], *etc*. For ML methods, physical or non-physical features that can indicate battery aging evolution are extracted manually or automatically (for deep learning) from the original voltage, current or temperature data for life prediction. These features can be classified into two typical categories: the features indicating the battery present health state, called 'state feature' for simplicity, and the features indicating battery degradation rate, called 'rate feature' for simplicity. The state features reflect the battery health from a multi-dimension space while the rate features capture the battery aging rate during a specific period of usage time. In the literature, the developed ML models generally predict battery EOL based on only the state features or the rate features.

For example, Severson and Attia et al. [17, 18] extract features from the first 100 cycles to indicate the evolution rates of battery aging under different operating conditions, and the mapping of these different rate features to the corresponding battery EOLs are learned through ML. To improve the application capability in practice, Zhang et al. [19] proposes to extract features from cycles within a moving window, so that the latest aging information of different battery cells under recent working conditions can be captured. The accurate life prediction of the rate feature-based methods require a similar initial health state of the battery to calculate the degradation rate. In [20], Attia et al. show that if the test batteries undergo longer calendar aging than the training batteries before experimenting, the battery lifetimes are generally overestimated with a mean value of 145 cycles, which phenomenon is due to the additional increase of the solid electrolyte interphase of the test battery, thus causing different initial health states between the training and test samples. However, no effective features are used to capture and update the battery health state in those studies [17-20].

Some researchers choose to use state features to predict battery RUL. For example, Zhang et al. [21] uses multi-dimension features extracted from the electrochemical impedance spectroscopy to indicate battery aging state, and in ref. [22], the authors use features of an equivalent circuit model (ECM) to predict battery RUL. Although accurate battery life is generally predicted in these papers [22, 23], the methods still suffer from mixed working conditions, as shown in [22]. Indeed, even the present health state is the same, the RUL of different battery cells can still be quite different owing to the cell-to-cell inhomogeneities in both aging characteristics and working conditions. To address this issue, ref. [23], respectively, extracts rate features from a moving window with a size of 30 cycles and extracts state features from the voltage differences between the present cycle and the initial cycle, which still suffers from the requirement that the initial battery state of health should be the same among different battery cells. Besides using the rate features extracted from the first 100 cycles as presented in [17], ref. [24] also extracts many physical features to indicate the present battery SOH. Although the early prediction capability of battery lifetime is improved, the rate features corresponding to the initial cycle cannot capture the changes in battery aging caused by varying working conditions [24].

To bridge the research gap, this study makes three contributions to realize accurate in-situ prediction of battery lifetime. First, the two types of features, *i.e.*, state features and rate features, are both involved in for improved battery life prediction and classification. The state features are six physical features of an ECM extracted from the voltage relaxation data, which data is easily collected onboard through the battery management system in practice. And the rate features are the differences between the relaxed voltage curves within a moving window (for life prediction), or the differences between the capacity vs. voltage curves at adjacent cycles (for life classification). Second, two ML models, which are, respectively, constructed based on Gaussian processes for life prediction and classification, are used to describe the relationships between the features and battery EOL. The feasibility of the method is evaluated systematically with different window sizes and voltage relaxation times. Finally, the method is validated with the aging data of 74 cells of three battery types. Experimental results show that the method predicts more accurate battery life than the benchmark based on only state features, with the accuracy improved by up to 67.09%. And a quick triple-classification of battery life is realized based on only two adjacent cycles' data with an overall accuracy larger than 90%.

The rest of this paper is organized as follows. Section II describes the experimental data, followed by the methodologies of feature engineering and ML developed in Section III, and the results of in-situ prediction and classification are presented in Section IV. Section V concludes the paper.

## II. DATA GENERATION

The battery cycling datasets generated by Zhu et al. [25] are used in this study. The dataset includes three types of batteries, *i.e.*, the NCA battery with a positive electrode of $LiNi_{0.86}Co_{0.11}Al_{0.03}O_2$, the NCM battery with a positive electrode of $LiNi_{0.83}Co_{0.11}Mn_{0.07}O_2$, and the NCM+NCA battery with a composite positive electrode of 42 (3) wt.% $Li(NiCoMn)O_2$ and 58 (3) wt.% $Li(NiCoAl)O_2$. The negative electrode of the NCA and NCM batteries mainly consists of 97 wt% graphite and 2 wt% silicon, while that of the NCM+NCA battery mainly consists of graphite. The batteries were cycled in a temperature-controlled chamber with different charge and discharge current rates. And the cycling conditions and generated dataset are summarized in Table I. Note that 1C is 3.5A for the NCA and NCM batteries and 2.5A for the NCM+NCA battery.

The NCA and NCM batteries were tested at three temperatures of 25°C, 35°C, and 45°C, while the NCM+NCA battery was only tested at 25°C. The charge current rate was 0.5 C for all battery cells except those NCA cells tested at 25°C where the charge rate of 0.25C was also loaded. The discharge rate was 1C for the NCA and NCM batteries, while for the NCM+NCA battery, three discharge rates of 1C, 2C and 4C were, respectively, loaded. The battery cells were charged with constant current (CC)-constant voltage (CV) with a cutoff voltage of 4.2V and a cutoff current of 0.05C, and discharged with CC with a cutoff voltage of 2.65V for NCA or 2.5V for NCM and NCM+NCA.

At the end of each charge or discharge, the rest time was set to 30 min for the NCA and NCM, while that for NCM+NCA was 60 min. The voltage relaxation data during battery resting was sampled every 2 min for the NCA and NCM and every 30 s for the NCM+NCA. For ease of presentation, CYX-Y/Z was used to name the operating conditions of cells, where X means the temperature, and Y/Z represents the charge/discharge current rate. The battery capacity fading trajectories for the three datasets are illustrated in Appendix Fig. 1. It shows that the battery capacity generally degrades quite differently under different operating conditions, as observed in practice. And even under the same operating condition, the capacity degradation shows high cell-to-cell variations, which induces great difficulty for online battery end-of-life (EOL) prediction. Detailed descriptions of the data and experimental conditions can be found at https://doi.org/10.5281/zenodo.6379165.

TABLE I
CYCLED BATTERIES AND CYCLING CONDITIONS

| Datasets | Nominal capacity (Ah) | Lower cutoff voltage (V)-upper cutoff voltage (V) | Cycling temperature ($\pm 0.2$°C) | Charge current rate (C)/discharge rate (C) | Average end-of-life (cycles)/standard deviation (cycles) | Number of training/test cells |
|---|---|---|---|---|---|---|
| NCA | 3.5 | 2.65-4.2 | 25 | 0.25/1 | 286/43 | 2/3 |
|  |  |  | 25 | 0.5/1 | 168/18 | 6/7 |
|  |  |  | 35 |  | 501/29 | 1/1 |
|  |  |  | 45 |  | 480/103 | 11/10 |
| NCM | 3.5 | 2.5-4.2 | 25 | 0.5/1 | 353/65 | 2/2 |
|  |  |  | 35 |  | 982/41 | 2/2 |
|  |  |  | 45 |  | 617/20 | 4/12 |
| NCM+NCA | 2.5 | 2.5-4.2 | 25 | 0.5/1 | 485/12 | 1/2 |
|  |  |  |  | 0.5/2 | 481/22 | 1/2 |
|  |  |  |  | 0.5/4 | 480/10 | 1/2 |

### III. METHODOLOGY

*A. In-situ battery life prediction*

Fig. 1 displays the framework of in-situ battery life prediction under varying operating conditions. Fig. 1(a) shows that the batteries follow varying degradation trajectories amid mixed operating conditions. And the battery RUL is not only dependent on the current SOH, but also dependent on the performance degradation rate, which is determined by the future working condition. Therefore, to predict accurate battery life, it is necessary to capture both the present state and evolution rate of battery aging, as illustrated in Fig. 1(b). The detailed feature engineering process is shown in the following part. Fig. 1(c) shows that an ML model based on GPR is constructed to learn the relationship between the extracted features and battery aging, with the in-situ battery life prediction presented in Fig. 1(d).

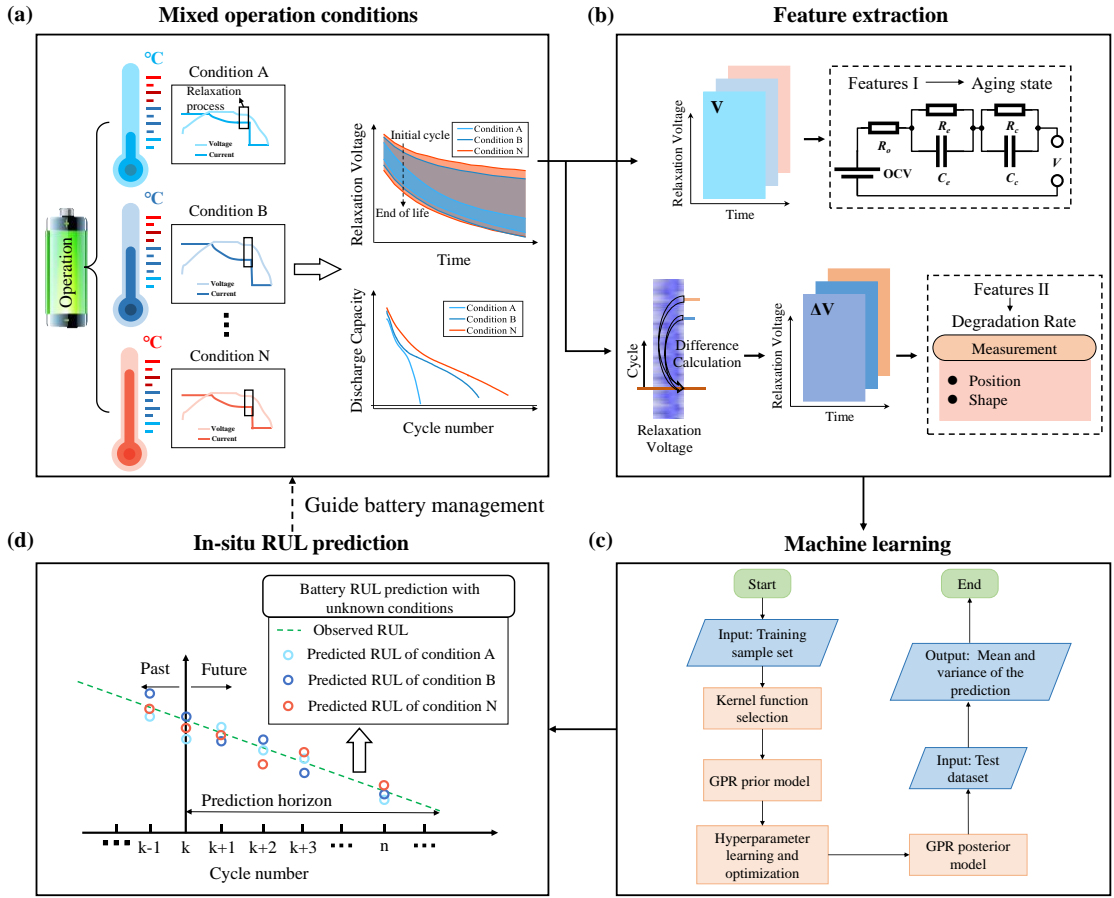

**Fig. 1: Framework of battery life prediction methodology under varying working conditions.** (a) Batteries working under different conditions. (b) Features extraction including the current state and evolution rate of battery aging. (c) GPR-based Machine learning methods. (d) In-situ battery life prediction.

**Feature Engineering for battery life prediction.** The voltage relaxation data after battery charging, which is easily collected in practice, is used for feature extraction in this study. As indicated in Fig. 1(a), the voltage relaxation curve moves down as battery degrades. Therefore, the single voltage relaxation curve indicates battery aging state and the difference between these two curves indicates battery degradation rate. According to ref. [22], the six parameters of a second-order ECM, which are identified based on the voltage relaxation data, inform physical aging information of batteries. For this reason, these six parameters are used as physical features to indicate battery aging state in this study (Fig. 1(b)).

According to ref. [22, 26], the battery relaxation voltage of a second-order ECM after charging is expressed as

$$U(t) = \text{OCV} - i(t)R_o - IR_e e^{-\frac{t}{R_e C_e}} - IR_c e^{-\frac{t}{R_c C_c}}, \tag{1}$$

where $U$ is the terminal (relaxation) voltage, $t$ is the sampling time, and $I$ is the cutoff current at the end of CV charging. The six parameters, *i.e.*, OCV is the open circuit voltage, $R_o$ is the ohmic resistance, $R_e$ and $C_e$ represent the resistance and capacitance of electrochemical polarization, and $R_c$ and $C_c$ represent the resistance and capacitance of concentration polarization. Note that $i(t) = I$ when $t = 0$, and $i(t) = 0$ when $t > 0$. The five parameters except $R_o$ are identified using a nonlinear least squared method to fit eq. (1) to the voltage data at $t > 0$. And then the resistance $R_o$ is identified at $t = 0$ as the following:

$$R_o = \frac{|V(0) - \text{OCV}|}{I} - R_e - R_c. \tag{2}$$

To capture the latest degradation rate of battery performance, the relaxed voltage difference within a moving window is extracted. Assuming there is a window covering battery degradation from cycle $n$ to cycle $m$, the voltage difference within this window is calculated as

$$\Delta V_{m-n}(t) = V_m(t) - V_n(t). \tag{3}$$

In this study, two statistical features, *i.e.*, the summation and the last value of $\Delta V_{m-n}(t)$, are extracted to measure the change of the relaxed voltage. These two features are, respectively, equal to $\sum_{t=\Delta T}^{T} \Delta V_{m-n}(t)$ and $\Delta V_{m-n}(T)$, where $\Delta T$ is the sampling interval and $T$ is the last sampling moment of voltage relaxation. Both of the two features evolve monotonously as the battery degrades, as shown in the following section.

**Machine learning regression algorithm.** Gaussian process regression (GPR) is a technique for non-parametric regression based on Gaussian processes (Fig. 1(c)), which defines a probability distribution over functions [19, 27] as

$$f(x) \sim GP(m(x), \kappa(x, x')), \tag{4}$$

where $m(x)$ and $\kappa(x, x')$ are the mean and the covariance functions, respectively:

$$m(x) = \mathbb{E}[f(x)], \tag{5}$$

$$\kappa(x, x') = \mathbb{E}\left[(f(x) - m(x))(f(x') - m(x'))^T\right]. \tag{6}$$

For the input data $X = (x_1, x_2, \ldots, x_n)_{d \times n}$ and output data $y = (y_1, y_2, \ldots, y_n)$, based on GPs, the joint probability distribution $p(f(x_1), f(x_2), \ldots, f(x_n))$ can be obtained with a mean of $m(x)$ and a covariance of $K(x)$ defined by $\kappa(x_i, x_j)$. Then the predictions are made at test indices $X^*$ by computing the conditional distribution $p(y^*|X^*, X, y)$, and the results are calculated as

$$p(y^*|X^*, X, y) = \mathcal{N}(y^*|m^*, \sigma^*), \tag{7}$$

where

$$m^* = K(X, X^*)^T K(X, X)^{-1} y, \tag{8}$$

$$\sigma^* = K(X^*, X^*) - K(X, X^*)^T K(X, X)^{-1} K(X, X^*). \tag{9}$$

In this paper, the automatic relevance determination (ARD)-based exponential covariance function is used:

$$\kappa_{EXP}(\pmb{x_i}, \pmb{x_j}) = \sigma_f^2 \exp\left(-\sqrt{\sum_{m=1}^{d} \frac{(x_{im} - x_{jm})^2}{l_m^2}}\right), \quad (10)$$

where $\sigma_f$ is the signal standard deviation, and $l_m$ is the length scale of the $m$th feature. The relative importance $w_m$ for the $m$th feature is calculated as

$$w_m = \frac{l_m}{\|l\|_1}. \quad (11)$$

The model performance is evaluated by using two evaluation metrics including the root-mean-square error (RMSE) and mean absolute percentage error (MAPE), which is calculated as

$$\text{RMSE} = \sqrt{\frac{1}{n}\sum_{i=1}^{n}(y_i - \hat{y}_i)^2}, \quad (12)$$

$$\text{MAPE} = \frac{1}{n}\sum_{i=1}^{n} \frac{|y_i - \hat{y}_i|}{y} \times 100\%, \quad (13)$$

where $y_i$ is the observed RUL, $\hat{y}_i$ is the predicted RUL, $y$ is the observed cycle of battery EOL, and $n$ is the total number of data units.

### B. Onboard battery life classification

Fig. 2 shows the methodology of in-situ battery life classification. In this scenario, the battery should be tested under a unified (standard) charging/discharging protocol, *e.g.*, CCCV charging/CC discharging, so that the battery RUL can be evaluated equally. Under this standard protocol, besides the voltage relaxation data, the voltage curve under CC charging or discharging can also be collected. In this case, the voltage difference vs. battery capacity ($\Delta Q(V)$) under CC charging/discharging can be used to capture the battery degradation rate, as presented in ref. [17], and the current state of battery aging is still indicated by the ECM features. Note that the test can be conducted either onboard or offboard for battery regrouping before the echelon utilization. Based on the test data, a GP-based classifier is constructed to classify the batteries into three groups, *i.e.*, short-, medium-, and long-lifetime.

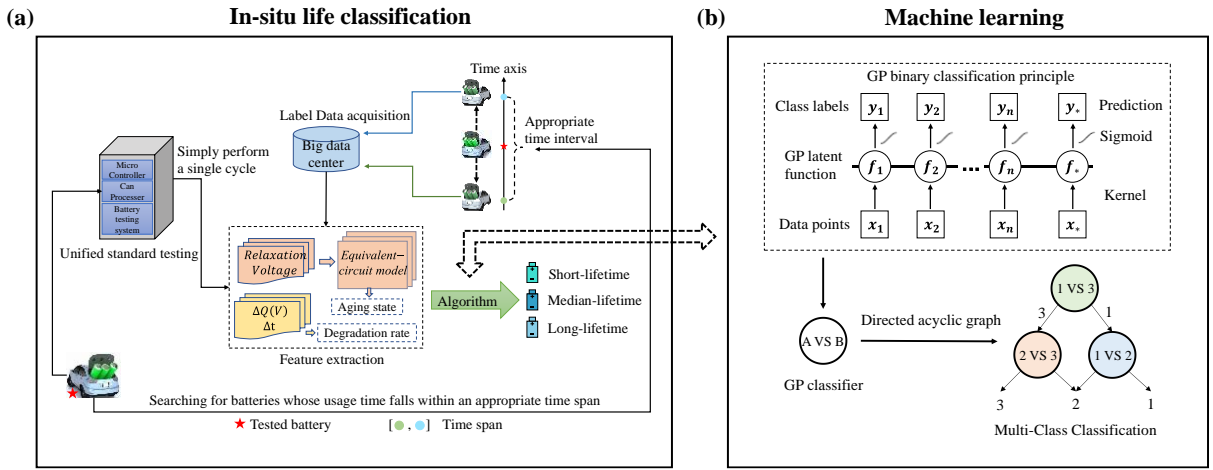

**Fig. 2: Scheme of battery life classification methodology with unified charging/discharging protocols.** (a) In-situ classification process. (b) Muti-classification machine learning method.

**Feature Engineering for life classification.** In this case, the ECM features are still used to indicate the battery aging state. And the battery degradation rate is captured by extracting features from the CC discharging data. To reduce the test cost, the characteristic differences of discharging data between two adjacent cycles are extracted. In other words, to realize efficient life classification the batteries are required to be tested for only two cycles for feature extraction. Two types of features are extracted to indicate the evolution rate of battery performance, *i.e.*, the variance of $\Delta Q_{m-n}(V)$ and the value of $\Delta t_{m-n}$. The feature $\Delta Q_{m-n}(V)$ represents the difference of discharge voltage vs. discharge capacity, which is expressed as

$$\Delta Q_{m-n}(V) = Q_m(V) - Q_n(V). \quad (14)$$

The variance of this signal, $Var(\Delta Q_{m-n}(V))$, contains rich information about battery electrochemical evolution, as presented in [17]:

$$Var(\Delta Q_{m-n}(V)) = \log_{10}\left(\left|\frac{1}{p-1}\sum_{i=1}^{p}(\Delta Q_i(V) - \overline{\Delta Q}(V))^2\right|\right), \quad (15)$$

where $p$ indicates the number of sampling points and is set to 1000 herein. The feature $\Delta t_{m-n}$ represents the difference of discharge time, which is expressed as [15]

$$\Delta t_{m-n} = \log_{10}(|t_m - t_n|), \quad (16)$$

where $t_m$ and $t_n$ are the discharge times of the cycles $m$ and $n$, respectively. Note that $m - n = 1$ for battery life classification.

**Machine learning classification algorithm.** As shown in Fig. 2(b), the task of dividing batteries into three groups is achieved by taking two steps to direct an acyclic graph based on a binary GP classifier [28, 29]. First, the classification is made through short- vs long-lifetime classifier. Second, if it is judged as long-lifetime, it enters the long- vs medium-lifetime classifier, otherwise it enters to the short- vs medium-lifetime classifier, and finally get the multi-classification result.

For each binary GP classifier, given input $X$ and their corresponding class labels $\mathbf{y} = [y_1, ..., y_n]^T \in \{-1, +1\}$, the class membership probability for a new test point $\mathbf{x}^*$ can be obtained by using a latent function $f$, whose value is then mapped into a domain of $[0,1]$ through a sigmoid function $\sigma(\cdot)$. Hence, the first is to calculate the distribution of the latent variable $f^*$ corresponding to $\mathbf{x}^*$:

$$p(f^*|X, \mathbf{y}, \mathbf{x}^*) = \int p(f^*|X, \mathbf{x}^*, \mathbf{f}) p(\mathbf{f}|X, \mathbf{y}) d\mathbf{f}, \quad (17)$$

where $\mathbf{f} = [f_1, ..., f_n]^T$, and then the predictive membership probability is obtained by

$$p(y^*|X, \mathbf{y}, \mathbf{x}^*) = \int \sigma(f^*) p(f^*|X, \mathbf{y}, \mathbf{x}^*) df^*. \quad (18)$$

The threshold for determining the battery lifetime, *i.e.*, long/medium/short, is determined dynamically at different battery SOHs as

$$N_{thr} = N_{thr}^{SOH=1} * \frac{SOH - 0.8}{0.2}, SOH > 0.8, \quad (19)$$

where the upper threshold $N_{thr\_upper}^{SOH=1} = 450$ and the lower threshold $N_{thr\_lower}^{SOH=1} = 180$ are set for the NCA battery dataset, and $N_{thr\_upper}^{SOH=1} = 800$ and $N_{thr\_lower}^{SOH=1} = 200$ are set for the NCM battery dataset.

## IV. RESULTS

*A. Features Extraction*

This section mainly presents the capability of the newly developed features in this study, *i.e.*, $\sum \Delta V_{m-1}(t)$ and $\Delta V_{m-1}(T)$ extracted from the differences of relaxed voltage curves, to inform battery degradation rate. The life indication capability of other features such as the ECM features, the variance of $\Delta Q_{m-n}(V)$, and the $\Delta t_{m-n}$ is referred to ref. [22], ref. [17], and ref. [15], respectively.

Fig. 3 shows the extracted features from the relaxed voltage curves. The current and voltage data of one complete cycle is displayed in Fig. 3(a), and the voltage relaxation curves after charging (stage III), as presented in Fig. 3(b), are used for features extraction. It shows that the relaxed voltage curves move downward as the battery ages, indicating the high potential for informing battery aging states. Appendix Fig. 2 shows the six physical features extracted from these voltage curves by using ECM, and the results show strong monotonous relationships between each physical feature and battery RUL. However, these monotonous relationships generally vary under different operating conditions, especially for the NCA and NCM batteries. For example, even at the same OCV the RUL is generally not the same for batteries under different cycling conditions. And even for batteries cycled with the same protocol, the feature with the same value ($OCV = 4.15\ V, e.g.$) also indicates different RULs owing to the cell-to-cell heterogeneity. These results indicate that only using the ECM features to predict battery RUL might not be enough. Therefore, physical features able to inform battery degradation rate are also required to further improve the battery RUL prediction performance.

Fig. 3(c) shows the differences of relaxed voltage curves between cycles $m$ and 1, where it shows that the curve differences also move downward as the cycle number increases. This shift of voltage difference curves indicates that a larger degradation rate of the battery lowers the $\Delta V_{m-1}(t)$ value. Two features including $\sum \Delta V_{m-1}(t)$ and $\Delta V_{m-1}(T)$ are further extracted, with the results of different types of batteries presented in Figs. 3(d-i). It shows that these two features both decrease monotonously as the battery ages for all types of batteries under different working conditions, which lays a solid foundation for effective ML-based battery RUL prediction. Note that the features $\sum Ah$ which accumulates all the ampere-hours through charging/discharging and $\sum t$ which indicates the calendar days are also introduced. The combination of these two features with the ECM features, which obtains the best battery RUL prediction in ref. [22], is used as a benchmark to evaluate the performance of the novel features developed in this study.

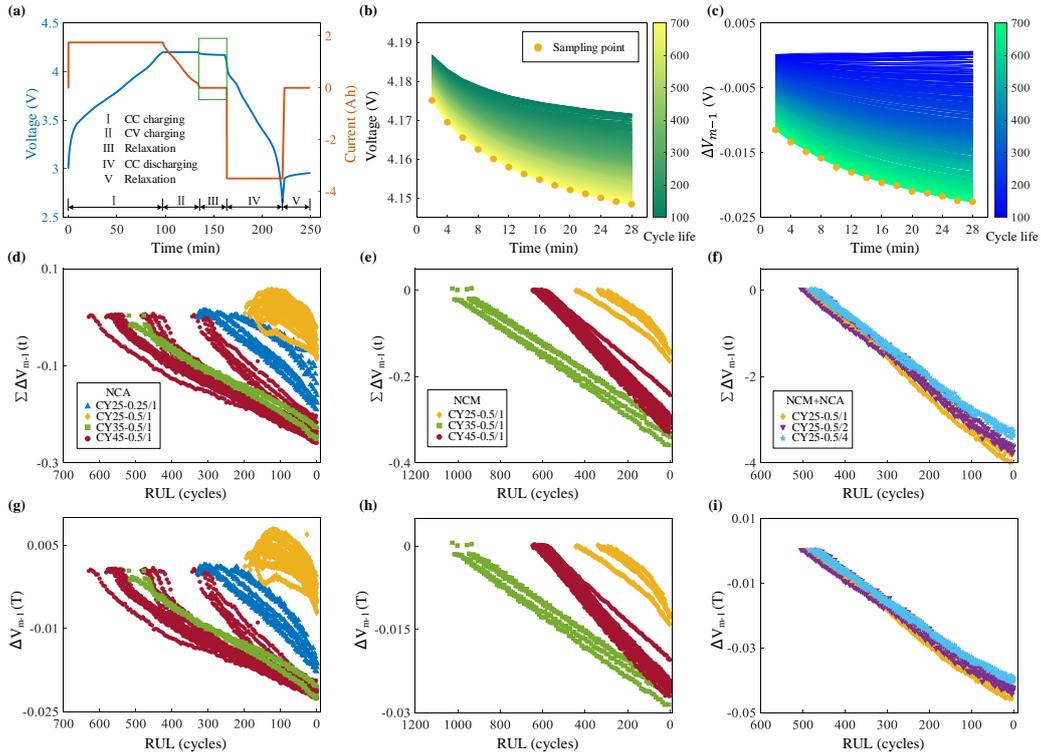

**Fig. 3: Features extraction from relaxed voltage curve of the battery.** (a) Voltage and current profiles of one complete cycle. (b) Voltage relaxation curves after battery full charging. (c) Differences of the voltage relaxation curves between cycles $m$ and 1. The feature $\sum \Delta V_{m-1}(t)$ of (d) NCA battery, (e) NCM battery, and (f) NCM+NCA battery. The feature $\Delta V_{m-1}(T)$ of (g) NCA battery, (h) NCM battery, and (i) NCM+NCA battery.

## B. Performance of in-situ battery RUL prediction

Fig. 4 shows the online RUL prediction results of the three battery datasets with mixed operating conditions. Figs. 4(a)-(c), respectively, show the relative importance of the extracted features based on the NCA, NCM, and NCM+NCA training datasets. For the NCA and NCM batteries, the two features of $\Delta V_{m-1}$ make important contribution to predicting battery RUL, with $\sum \Delta V_{m-1}(t)$ ranked the third for both NCA and NCM, and $\Delta V_{m-1}(T)$ ranked the second for NCM in the relative importance. For the NCA+NCM battery, because the variations of ECM features show high uniformity even against different working conditions (Appendix Fig. 2), it is already enough to predict accurate battery RUL by using the ECM features only. And it is unnecessary to introduce the features of $\Delta V_{m-1}$ to capture the battery degradation rate in this case. For this reason, $\sum \Delta V_{m-1}(t)$ and $\Delta V_{m-1}(T)$ both contribute little to predicting the RUL of NCM+NCA.

Figs. 4(d)-(i) present the results of predicted RULs versus observed RULs, where the insets show the distribution histograms of prediction errors. It is observed that the predicted RULs vary around the observed values, with the prediction errors close to zero for different types of batteries under different cycling conditions. TABLE II lists the RMSEs of battery RUL prediction with different types of features, where STATS indicates the eight statistical features, *i.e.*, the variance, skewness, kurtosis, max,

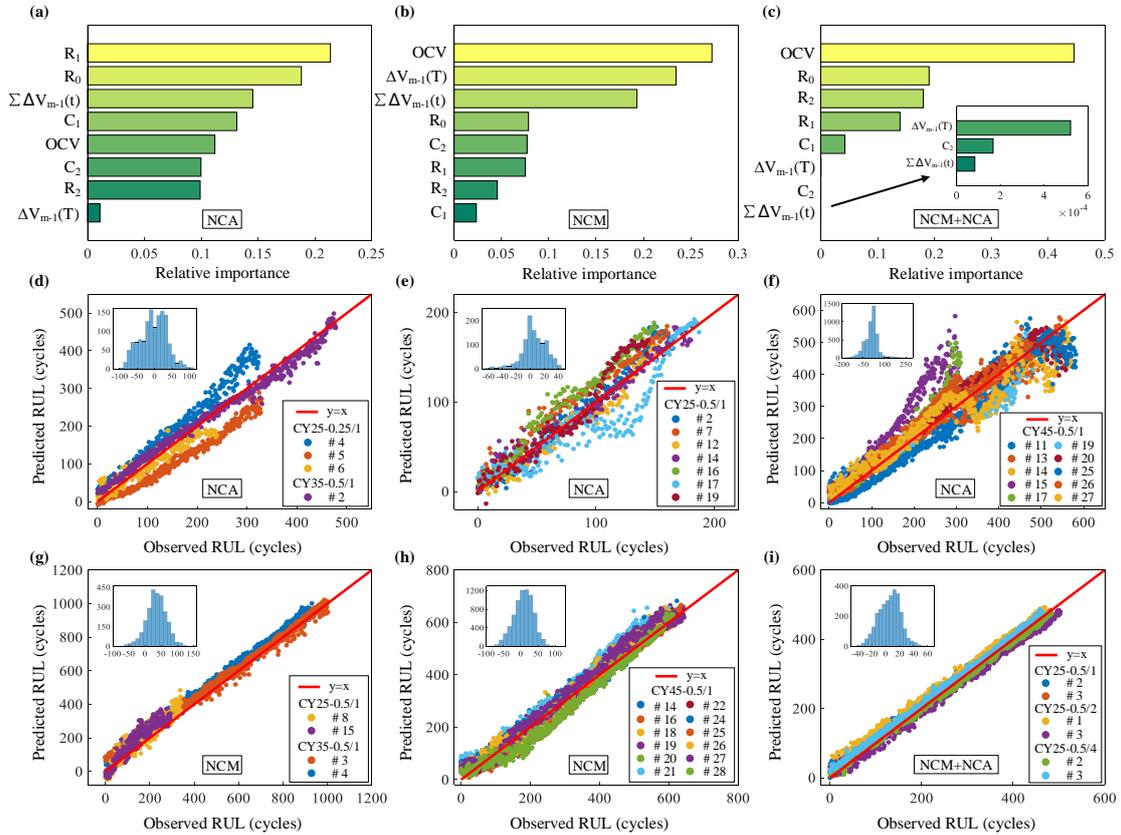

**Fig. 4: RUL prediction results with unknown working conditions.** Relative importance of features based on the training dataset of the (a) NCA battery, (b) NCM battery, (c) NCM+NCA battery. Prediction results of the NCA battery under cycling conditions of (d) CY25-0.25/1 and CY35-0.5/1, (e) CY25-0.5/1, (f) CY45-0.5/1. Prediction results of the NCM battery under cycling conditions of (g) CY25-0.5/1 and CY35-0.5/1, (h) CY45-0.5/1. Prediction results of the NCM+NCA battery under cycling conditions of (i) CY25-0.5/1, CY25-0.5/2, and CY25-0.5/4.

mean, mean, $\sum \Delta V_{m-1}(t)$, and $\Delta V_{m-1}(T)$ extracted from the curve $\Delta V_{m-1}(t)$. The feature ECM is used to capture the current aging state, while the feature STATS is used to capture the degradation rate of battery performance under different cycling conditions.

It shows that the ML model based on ECM or STATS generally predicts RUL with larger errors than those based on other features, indicating that only capturing one state (present state or evolution rate) of battery aging is not enough to predict accurate RUL. And the model based on STATS even fails to predict battery RUL in some cases especially for NCM, with an RMSE up to 317.6 cycles. The prediction performance is highly improved when both the present state and evolution rate of battery aging are captured by the novel feature ECM+$\sum \Delta V_{m-1}(t)+\Delta V_{m-1}(T)$. The model with this newly developed feature predicts the most accurate RUL for NCA and NCM under different cycling conditions, with the RMSEs generally less than 50 cycles. The prediction results of NCM+NCA are quite different from those of other batteries, where the prediction errors based on ECM, STATS and the novel features are similar to each other. This prediction similarity is caused by the high cell-to-cell uniformity of the NCM+NCA batteries, where the ECM features (*e.g.*, OCV and $R_o$ shown in Appendix Fig. 2) and $\Delta V_{m-1}(t)$ features (*e.g.*, $\sum \Delta V_{m-1}(t)$ and $\Delta V_{m-1}(T)$ shown in Fig. 3) under different working conditions change similarly as the batteries age. In this case, the battery RUL can be predicted accurately by capturing either state of battery aging. The model based on novel features also normally performs better than the benchmark model except for NCM+NCA, with the accuracy improvement up to 67.09%.

Note that for the benchmark, although the feature $\sum Ah$ can be collected easily in practice, the accumulation errors of the onboard current sensors can cause large calculation errors to $\sum Ah$. Besides, in the experiment the batteries are fully charged and

TABLE II

RMSE OF GPR-BASED BATTERY RUL PREDICTION WITH DIFFERENT TYPES OF FEATURES

| Battery | Cycling conditions | ECM | STATS | Benchmark: ECM+$\sum Ah$+$\sum t$ | ECM+$\sum \Delta V_{m-1}(t)$ +$\Delta V_{m-1}(T)$ | Improvement to benchmark |
|---|---|---|---|---|---|---|
| NCA | CY25-0.25/1 | 60.51 | 160.08 | 50.78 | 45.48 | 10.44% |
| | CY25-0.5/1 | 30.82 | 30.62 | 19.57 | 18.39 | 6.03% |
| | CY35-0.5/1 | 29.33 | 53.46 | 59.53 | 19.59 | 67.09% |
| | CY45-0.5/1 | 48.37 | 71.21 | 47.30 | 48.55 | -2.64% |
| NCM | CY25-0.5/1 | 72.03 | 317.60 | 47.45 | 53.59 | -12.94% |
| | CY35-0.5/1 | 88.32 | 161.10 | 52.45 | 45.11 | 13.99% |
| | CY45-0.5/1 | 63.87 | 77.37 | 40.60 | 29.75 | 26.72% |
| NCM+ NCA | CY25-0.5/1 | 12.27 | 10.78 | 7.62 | 12.27 | -61.02% |
| | CY25-0.5/2 | 22.04 | 25.61 | 16.59 | 22.03 | -32.79% |
| | CY25-0.5/4 | 13.29 | 13.51 | 15.43 | 13.92 | 9.79% |

discharged, and thus the accumulation $\sum Ah$ essentially contains the information of battery capacity degradation, which might be the main reason for the improved RUL prediction accuracy when being combined with ECM. While in practice, the batteries are usually partially charged and discharged, and $\sum Ah$ cannot reflect the battery capacity degradation. For these reasons, the benchmark model is expected to perform worse in practice than that in experiment as shown in Table II.

In practice, there will be scenarios when the early cycling data of battery is not available or the battery operating conditions in the future are different from those in the past. In these cases, it is important to capture the latest aging information of battery, and therefore, the RUL prediction performance of the developed methodology needs to be evaluated with different initial aging states. Fig. 5 depicts the in-situ RUL prediction results from different cycles of 1, 50, 100, and 150, where Figs. 5(a)-(c) show the prediction results of the three battery datasets, respectively. The histograms represent the RMSE and the broken lines represents the MAPE. The results show that the prediction performance of the ML model, which is constructed based on the developed features, does not reduce as the battery degrades. And the model still generally performs the best for NCA and NCM among the models with different features.

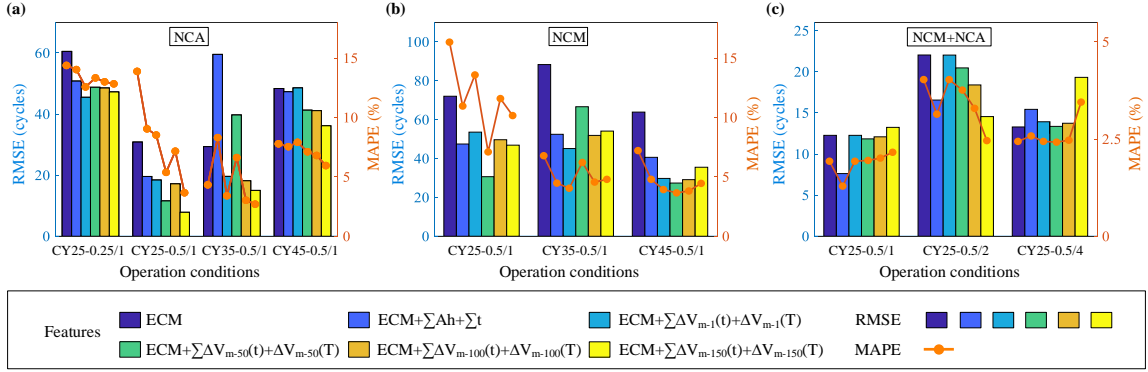

**Fig. 5: RUL prediction results from different aging states.** (a) The NCA battery. (b) The NCM battery. (c) The NCM+NCA battery.

To further evaluate the feasibility of the developed method, the battery RUL is predicted based on voltage data of different relaxation times, with the prediction results shown in Fig. 6, where the predicted RULs under different operating conditions of the same battery type are summarized and presented. Generally, a less relaxation time, which is required to collect the voltage data for accurate battery RUL prediction, indicates less energy dissipated by the BMS and brings less interruption to the normal usage of the power systems. Note that a minimum of six data points are required for the ECM parameter identification using nonlinear fitting technique. These six data points, respectively, account for 12 min for NCA and NCM owing to a sampling interval of 2 min, and 3 mins for NCM+NCA owing to a sampling interval of 30 s. Figs. 6(a-c) show that for different types of batteries, the developed methodology predicts stable and accurate battery RUL at different relaxation times, with a slight increase of the prediction errors when only six data points are used. The benchmark, however, shows unstable battery RUL prediction, especially for the NCM battery (Fig. 6(b)), when the prediction errors rise as the relaxation time decreases. Appendix Fig. 3 shows the RUL prediction results of batteries at different relaxation times and aging states. It shows that even from different starting cycles, the developed GPR model predicts accurate battery RUL when only six data points of the relaxed voltage are available.

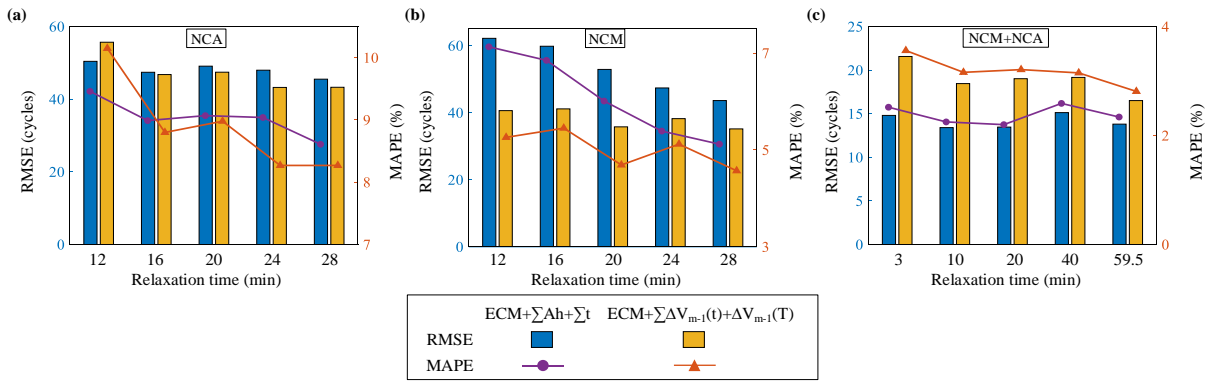

**Fig. 6: Battery RUL prediction results of all battery types at different relaxation times.** (a) NCA battery. (b) NCM battery. (c) NCM+NCA battery.

*C. Performance of onboard battery life classification*

The battery life classification before reusage is generally conducted after reaching similar retirement conditions, *e.g.*, the capacity fades to 80% of its initial value. In this case, which is different from the in-situ life prediction of battery, it is unreasonable to use the whole lifetime data of battery for training a life classifier, because the standard tests required herein are only conducted on batteries with similar aging stages. In this study, the batteries that are cycled similar times are considered to have similar aging stages. For example, at a test cycle of 500 and with a window size of 100 cycles to determine the similar aging stage, the battery aging data within [450, 550] cycles will be used to train the life classifier.

Fig. 7 displays the battery life classification results, where the battery lives are divided into three groups of short lifetimes, medium lifetimes, and long lifetimes. The method developed to dynamically determine the thresholds of long/medium/short lifetimes is referred to Section III. Note that because the lifespans and capacity decay trajectories of the NCM+NCA battery are very close to each other, the dataset is not used for life classification in this case. Figs. 7(a) and (c) show the battery life distribution at all cycles (data samples). It indicates that for NCA, the batteries tested at 25 °C are mainly classified to groups of short and medium lifetimes, while the batteries tested at 35 °C or higher are mainly classified to groups of long lifetimes. For NCM, the batteries tested at 35 °C are all with long lifetimes, while the batteries tested at higher or lower temperatures cover all three life spans. Figs. 7(b) and

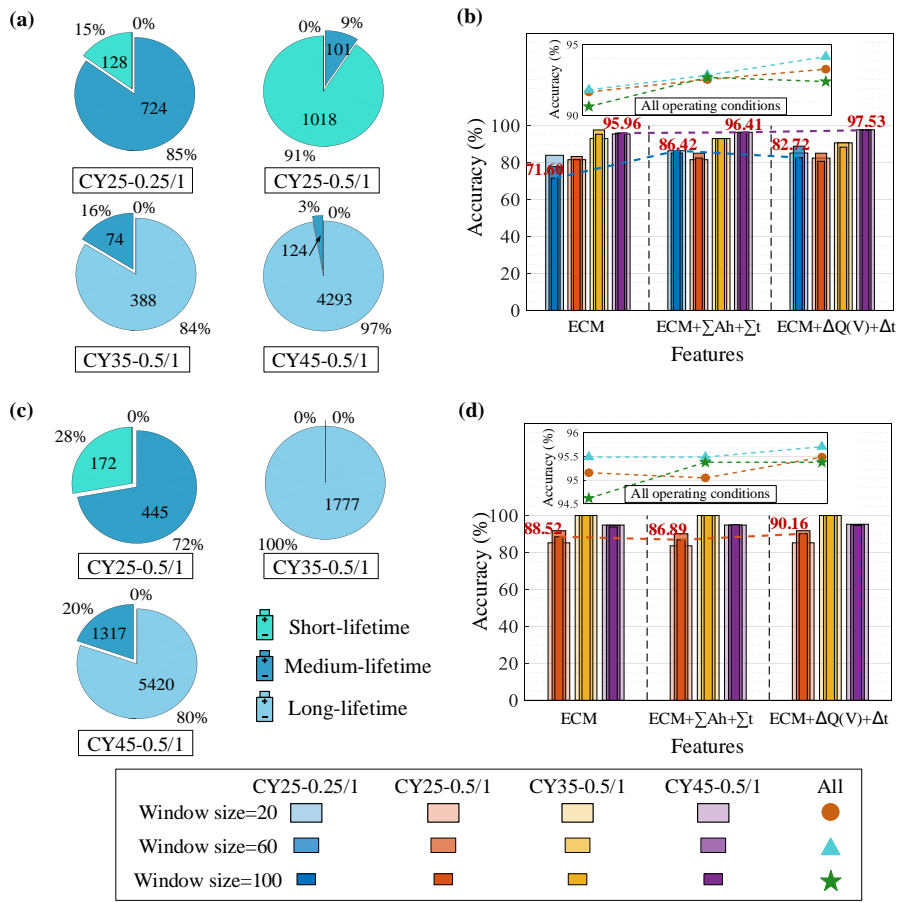

**Fig. 7: Battery life classification results.** The cycle life distribution of all samples of (a) the NCA battery and (c) the NCM battery. Life classification accuracies under different operating conditions of (b) the NCA battery and (d) the NCM battery.

(d) show that for different types of batteries, the developed methodology generally classifies batteries into the right groups with higher classification accuracy than ECM under each operating condition. The overall classification accuracy of the developed classifier is around 95%, which is comparable to the benchmark method. Besides, the classification results show limited variations with the window sizes of 100, 60 and 20 cycles under each operating condition, indicating the high robustness against the various battery aging states in practical classifications.

TABLE III lists the classification results of battery life based on different types of features, where at each test cycle, the battery aging data with a window size of 100 cycles is used to train the classifiers. Table III shows that for different battery types, the ECM classifier obtains high classification accuracies at different temperatures but the NCA battery at 25 °C, in which case the classification accuracy is low to 71.6%. This low accuracy indicate that besides the battery aging state, it is necessary to also include the degradation rate of battery for improved life classification. Indeed, the developed classifier capturing both the present state (ECM) and evolution rate ($var(\Delta Q(V))+\Delta t$) of battery aging improves the accuracy of 71.6% by more than 10% to 82.72%. Generally, table III shows that the developed method performs better than the classifiers based on ECM or $var(\Delta Q(V))+\Delta t$ under different operating conditions.

One exception exists for NCA at 35 °C, when the classification accuracy of the developed classifier is about 7% lower than that based on the ECM classifier. This lower accuracy is possibly due to the limited aging data collected under that condition, where only two battery cells were tested, and one of them was used for training and another one was used for test. Besides the ECM features, the developed classifier also includes the features of $var(\Delta Q(V_{m-n}))$ and $\Delta t_{m-n}$, which thus needs more data to construct a reliable relationship between the input and the output of the ML model. Overall, table III showcases that the developed method classifies batteries with comparable accuracy to that of the benchmark.

The life classification results of batteries at different SOHs are shown in Fig. 8, where Figs. 8(a)-(c) show the classified battery samples at SOHs of 100%, 95% and 85%, respectively. The vertical dashed lines of different colors represent the classification thresholds of lifetime for the NCA and NCM batteries, which are dynamically updated with the corresponding SOH. For each battery type, the dashed lines divide the battery into three groups of short-, medium- and long-lifetime.

TABLE III
CLASSIFICATION ACCURACY WITH DIFFERENT TYPES OF FEATURES

| Battery | Cycling conditions | ECM | $var(\Delta Q_{m-n}(V))+\Delta t_{m-n}$ | Benchmark: ECM+∑Ah+∑t | ECM+ $var(\Delta Q_{m-n}(V))+\Delta t_{m-n}$ | Improvement to benchmark |
|---|---|---|---|---|---|---|
| NCA | CY25-0.25/1 | 71.60 | 66.67 | 86.42 | 82.72 | -4.28% |
| | CY25-0.5/1 | 81.58 | 79.82 | 82.46 | 80.70 | -2.13% |
| | CY35-0.5/1 | 95.35 | 81.40 | 93.02 | 88.37 | -5.00% |
| | CY45-0.5/1 | 95.96 | 94.62 | 96.41 | 97.53 | 1.16% |
| | All | 90.64 | 88.01 | 92.69 | 92.40 | -0.31% |
| NCM | CY25-0.5/1 | 88.52 | 72.13 | 86.89 | 90.16 | 3.76% |
| | CY35-0.5/1 | 100 | 90.64 | 100 | 100 | 0.00% |
| | CY45-0.5/1 | 93.81 | 90.56 | 94.99 | 94.69 | -0.32% |
| | All | 94.62 | 89.34 | 95.38 | 95.38 | 0.00% |

The results show that generally, the classification accuracy increases as the battery degrades, especially for NCA. Fig. 8(a) illustrates that when fresh new NCA batteries are classified, the accuracy is low to 61.90%, which value increases to 90.48% at 95% SOH. And when the cycle lives are close to the EOL, only one sample is misclassified. This high classification accuracy near the battery EOL ensures the highly efficient and accurate regrouping of retired batteries based on the developed methodology. For NCM, the batteries are rightly classified throughout the whole lifetime, again validating the high classification performance of the method.

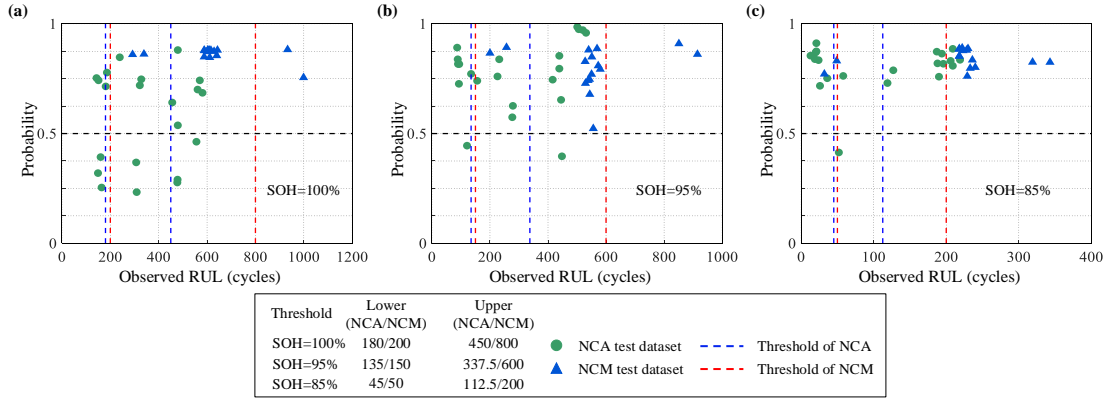

**Fig. 8: Life classification results of batteries at different SOHs.** (a) SOH=100%. (b) SOH=95%. (c) SOH=85%. Note that the *y*-axis indicates the probability that the observed battery sample is classified into the group of the corresponding lifetimes. Therefore, the samples above the 0.5 probability line represent the ones that are rightly classified, and vice versa.

## V. CONCLUSIONS

Battery life prediction, as a core function of the battery management system, informs the remaining useful life (RUL) for preventative maintenance of a power system. And the life classification provides important information for the efficient regrouping of large-scale retired battery cells. To further improve the current performance of battery life prediction and classification, this study develops a machine learning-based method by capturing both the present state and evolution rate of battery aging.

In this study, six physical features of an equivalent circuit model (ECM) are extracted from the voltage relaxation data to indicate the present aging state of the battery. In the case of battery life prediction, the differences between the voltage relaxation curves at different cycles, $\Delta V_{m-n}(t)$, are used to indicate the battery degradation rate through those cycles. While for battery life classification, the case is quite different, where the battery degradation rate is captured by the differences between the curves of capacity versus voltage between two adjacent cycles, indicated as $\Delta Q_{m-n}(V)$. Two machine learning models based on Gaussian Processes (GPs) are, respectively, established to describe the relationships between the physical features and battery lifetime for life prediction and classification purposes. Unlike the double-classification generally conducted in the literature, this study performs a triple-classification of battery lifetimes involving short, medium, and long.

The model performance is validated based on the aging data of 74 batteries of three types under different operating conditions. Experimental results show that, based on a minimum of 6 voltage data points (taking 3-12 min to sample), the method predicts battery RUL accurately with the root-mean-square errors (RMSEs) and mean absolute percentage errors (MAPEs) being, respectively, within 50 cycles and 10%. And the battery life can still be predicted accurately as the moving window moves forward to capture the latest aging information of the battery. Compared with the benchmark model that only captures the battery aging state, the developed method improves the life prediction accuracy by up to 67.09%. The battery life is classified accurately with the overall accuracies larger than 90% based on only two adjacent cycles' data, and the classification accuracy increases as the battery degrades, enabling efficient and accurate regrouping of the retired battery cells in practice. These results indicate the necessity to include both the present state and evolution rate of battery aging for improved life prediction and classification.

# APPENDIX

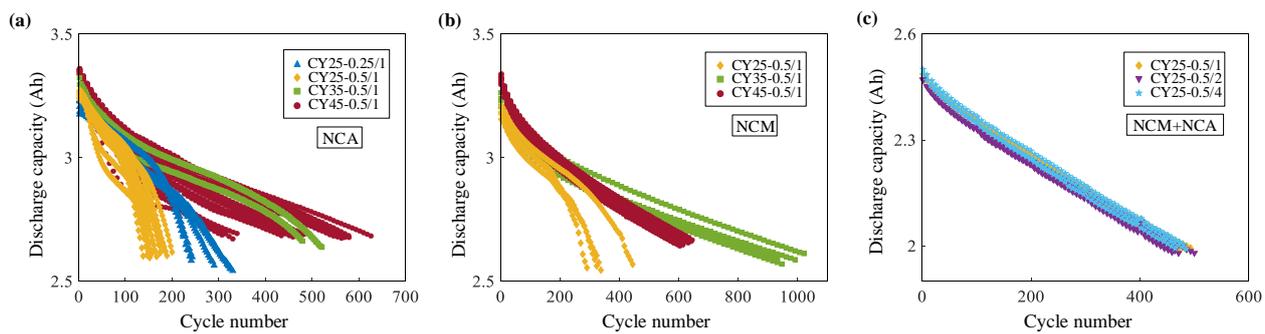

**Appendix Fig. 1: Capacity degradation versus cycle number.** (a) NCA battery. (b) NCM battery. (c) NCM+NCA battery.

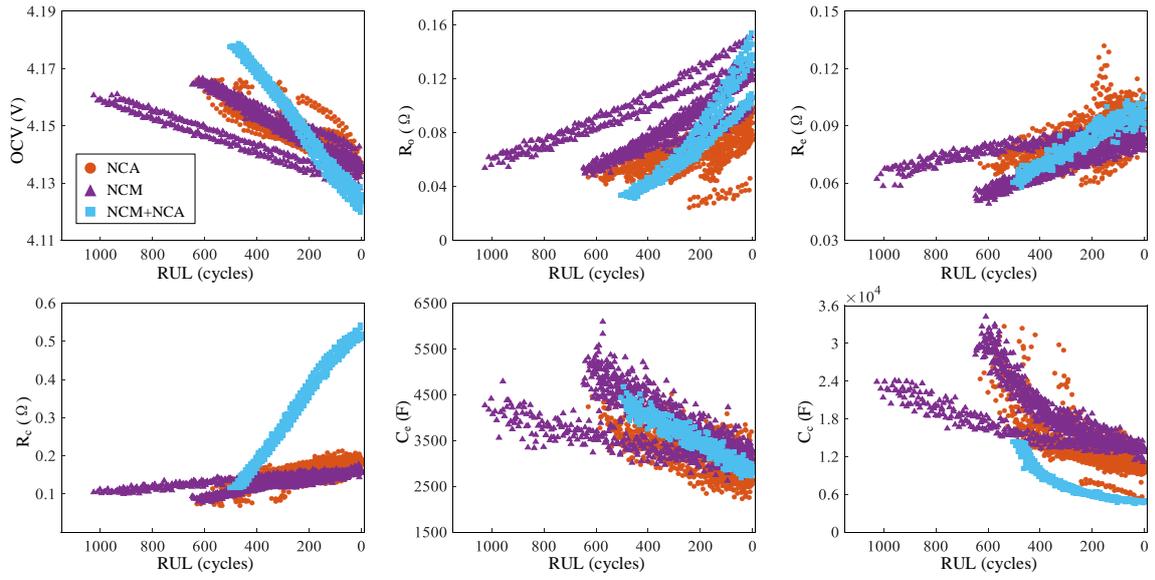

**Appendix Fig. 2: ECM features extracted from the voltage relaxation curves of different types of batteries.** (a) Open circuit voltage (OCV). (b) Ohmic resistance ($R_o$). (c) Electrochemical polarization resistance ($R_e$). (d) Concentration polarization resistance ($R_c$). (e) Electrochemical polarization capacitance ($C_e$). (f) Concentration polarization capacitance ($C_c$).

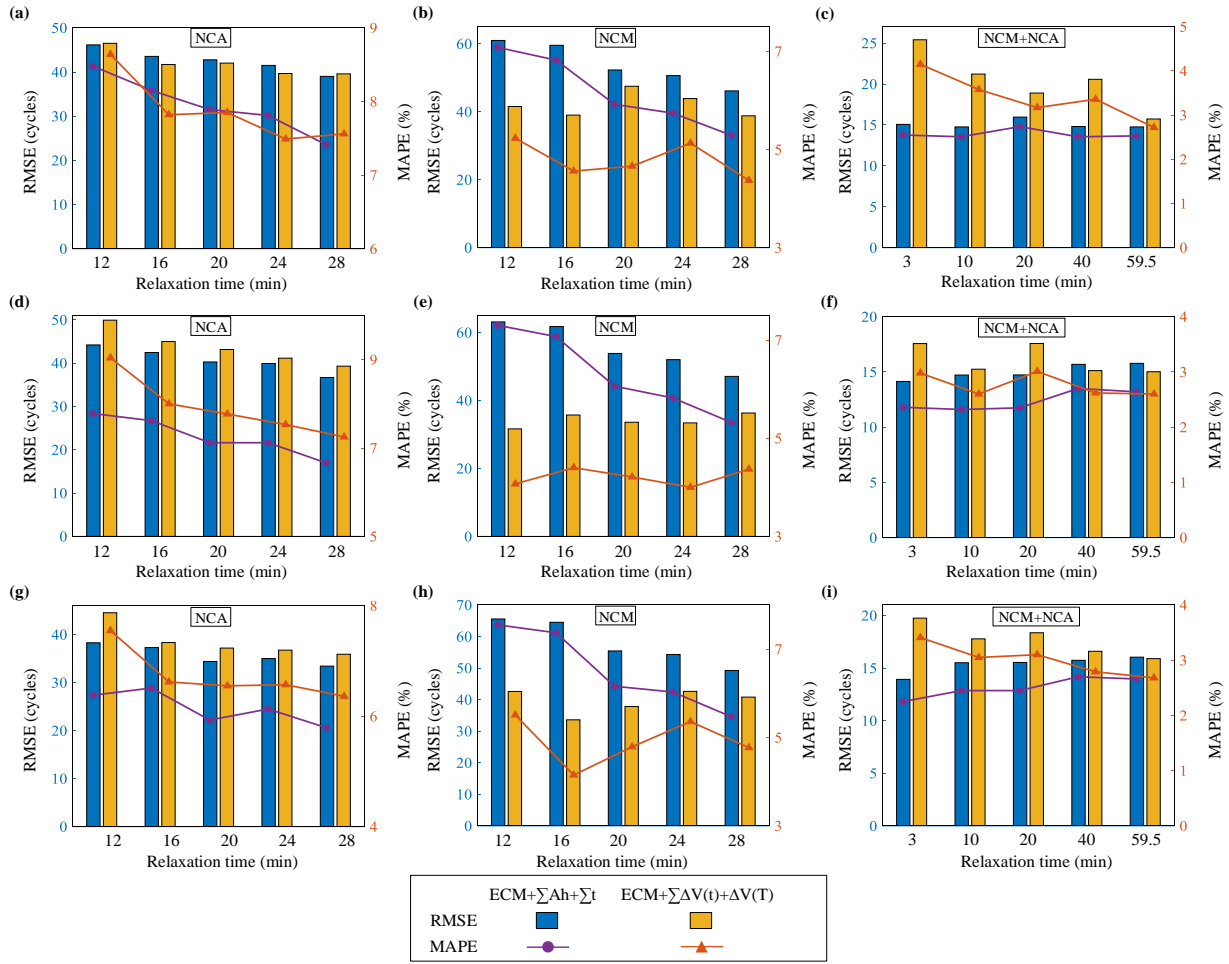

**Appendix Fig. 3: RUL prediction results at different relaxation times of all battery types.** Prediction results from 50 cycles of (a) NCA, (b) NCM, and (c) NCM+NCA. Prediction results from 100 cycles of (d) NCA, (e) NCM, and (f) NCM+NCA. Prediction results from 150 cycles of (g) NCA, (h) NCM, and (i) NCM+NCA.